\documentclass[journal]{IEEEtran}
\usepackage{comment}
\usepackage{adjustbox}
\usepackage{url}
\usepackage{makecell}
\usepackage{xcolor}
\usepackage{cite}
\usepackage{textcomp}
\usepackage{siunitx}
\usepackage{graphics}
\usepackage{multirow}
\usepackage{booktabs}
\usepackage{lscape}
\usepackage{graphicx}
\usepackage{verbatim}
\usepackage[hidelinks]{hyperref}
\usepackage{cite}
\usepackage[flushleft]{threeparttable}
\usepackage{multirow}
\usepackage[normalem]{ulem}
\useunder{\uline}{\ul}{}
\usepackage{subfig}
\usepackage{amsmath,amssymb,amsfonts}
\usepackage{algorithmic}
\usepackage{graphicx}
\usepackage{pgfplotstable}
\usepackage{textcomp}

\ifCLASSINFOpdf
\else
\fi
\begin{document}

\title{A Pilot Study on Visually Stimulated Cognitive Tasks for EEG-Based Dementia Recognition}
\author{Supavit~Kongwudhikunakorn,
        Suktipol~Kiatthaveephong,
        Kamonwan~Thanontip,
        Pitshaporn~Leelaarporn,
        Maytus~Piriyajitakonkij,
        Thananya~Charoenpattarawut,
        Phairot~Autthasan,
        Rattanaphon~Chaisaen,
        Pathitta~Dujada, Thapanun~Sudhawiyangkul,
        Vorapun~Senanarong$^{*}$
         and
        ~Theerawit~Wilaiprasitporn$^{*}$,~\IEEEmembership{Member,~IEEE}
\thanks{This work was supported by PTT Public Company Limited, The SCB Public Company Limited, Thailand Science Research and Innovation (SRI62W1501), The Office of the Permanent Secretary of the Ministry of Higher Education, Science, Research and Innovation, Thailand (RGNS63-252) and National Research Council of Thailand (N41A640131) \textit{(Supavit Kongwudhikunakorn and Suktipol Kiatthaveephong contributed equally to this work) ($^{*}$Corresponding author: Vorapun Senanarong, Theerawit Wilaiprasitporn).}}
\thanks{S.~Kongwudhikunakorn, S.~Kiatthaveephong, K.~Thanontip, P.~Leelaarporn, M.~Piriyajitakonkij, P.~Autthasan, R.~Chaisaen, T.~Sudhawiyangkul and T.~Wilaiprasitporn are with Bio-inspired Robotics and Neural Engineering (BRAIN) Lab, School of Information Science and Technology (IST), Vidyasirimedhi\break Institute of Science \& Technology (VISTEC), Rayong, Thailand (e-mail: theerawit.w@vistec.ac.th).}
\thanks{T.~Charoenpattarawut is with Bio and Brain Engineering Department, Korea Advanced Institute of Science and Technology (KAIST), Daejeon, South Korea}
\thanks{P.~Dujada and V.~Senanarong are with Department of Medicine, Faculty of Medicine Siriraj Hospital, Mahidol University, Bangkok, Thailand and Neurocognitive Disorders \& Neural Computing Research Group}
}

\maketitle

\begin{abstract}
In the status quo, dementia is yet to be cured. Precise diagnosis prior to the onset of the symptoms can prevent the rapid progression of the emerging cognitive impairment. Recent progress has shown that Electroencephalography (EEG) is the promising and cost-effective test to facilitate the detection of neurocognitive disorders. However, most of the existing works have been using only resting-state EEG. The efficiencies of EEG signals from various cognitive tasks, for dementia classification, have yet to be thoroughly investigated. In this study, we designed four cognitive tasks that engage different cognitive performances: attention, working memory, and executive function. We investigated these tasks by using statistical analysis on both time and frequency domains of EEG signals from three classes of human subjects: Dementia (DEM), Mild Cognitive Impairment (MCI), and Normal Control (NC). We also further evaluated the classification performances of two features extraction methods: Principal Component Analysis (PCA) and Filter Bank Common Spatial Pattern (FBCSP). We found that the working memory related tasks yielded good performances for dementia recognition in both cases using PCA and FBCSP. Moreover, FBCSP with features combination from four tasks revealed the best sensitivity of 0.87 and the specificity of 0.80. To our best knowledge, this is the first work that concurrently investigated several cognitive tasks for dementia recognition using both statistical analysis and classification scores. Our results yielded essential information to design and aid in conducting further experimental tasks to early diagnose dementia patients.

\end{abstract}


\section{Introduction}
As addressed by the World Health Organization (WHO), approximately 50 million people have dementia worldwide, with 10 million new cases reported each year. One of the leading causes of disability and dependency among the elderly is dementia. It does not only cause neurological impairments in the patients but also instigates a severe impact on their families and society as a whole \cite{dementia_who}. Dementia is a chronic syndrome induced by a gradual degradation and death of neurons. The most recognizable symptoms include the deterioration of cognitive capabilities such as memory, thinking, and the ability to perform daily living activities \cite{fwcfbsbbc2018,harrison_medicine}. Additionally, impairments in judgment, learning, executive functions, language and perception are symptoms of dementia \cite{wwywyd2015}. Even though no standard cure can revert the progression of dementia, it is possible to decelerate the cognitive deterioration if detected and treated in an early stage \cite{utmtyt2013}.

Numerous works have revealed that electroencephalography (EEG) is a cost-effective, portable, and non-invasive electrophysiological tool that accurately reflects the brain's activity \cite{cf2020, msbimmbm2019}.
EEG is used to record the detectable neural oscillations with varied frequency and amplitude, corresponding to many factors such as mental states (resting, thinking, memorizing), cognitive load, age, diseases, and mental tasks \cite{shlkw2020, ampst2020, sbayhk2020}.
It also yields a higher temporal resolution in comparison to the other neuroimaging techniques \cite{mcps2014}. Thus, it is a widely accepted method to help the diagnosis of neurodegenerative diseases \cite{mkmrrhsb2021,dkcokck2021}.
Most of the existing works have focused only on using resting-state EEG to classify neurological disorders of patients \cite{nyknkt2020, dzczmuvhlzbs2019, blwlyl2014, fwcfbsbbc2018}, but only few studies used EEG recorded when performing cognitive tasks\cite{hwjw2017, sharma2020iterative}.
\textcolor{black}{
Furthermore, the previous studies \cite{garn2014quantitative, garn2015quantitative} also reported that the cognitive task EEG provided the augmentative information compared to resting-state EEG.
As a result, this motivates us to further explore and design cognitive tasks to enhance MCI and dementia recognition.
}



In this work, four visual-based cognitive tasks--including Fixation, Mental Imagery, Symbol Recognition, and Visually Evoked Related Potential (VERP)--are proposed to extract distinguishable neural activity patterns from three different groups of human subjects: Normal Control (NC), Mild Cognitive Impairment (MCI) and Dementia (DEM).
We aim to identify the most suitable cognitive tasks for dementia detection, the methods to detect dementia using EEG signals.
Unlike previous works, this work studies dementia subjects instead of Alzheimer’s subjects \cite{mehjmnoswwa2018, fwcfbsbbc2018, msbimmbm2019} to produce more general models that can detect other types of dementia.
\textcolor{black}{
EEG signals are analyzed in both time and frequency domains, which have been utilized as a sensitive tool for cognitively impaired patients \cite{hscssk2018}.
}
A statistical test is used to indicate the differences in EEG signals among three classes of subjects.
Support Vector Machine (SVM) \cite{svm_paper} is then applied as a classifier with two feature extraction techniques: Principal Component Analysis (PCA) and Filter-Bank Common Spatial Pattern (FBCSP).
We find that FBCSP--one of the best feature extraction methods used in Brain-Computer Interfaces--provides promising result for dementia screening.

\textcolor{black}{
The major contributions of the current study are:
\begin{enumerate}
  \item Four visual-based cognitive tasks that captured various cognitive functions are designed and investigated to find the most suitable one for classifying the cognitive disorders.
  \item This is the first study that explores the effectiveness of FBCSP, one of the most powerful feature extraction methods in Brain-Computer Interfaces, in dementia classification.
  \item To increase the effectiveness of classification and decrease the level of weariness on subjects, we determine the suitable task and number of trials for each classification pair.
  \item To be more practical, this work evaluates the performance as patient-based classification rather than trial-based.
\end{enumerate}
}
The following sections are organized as follows: materials and methods, results, discussion, and conclusion.

\hfill 
\section{Materials and Methods}
\label{sec_methods}

Recruited subjects from three groups, namely DEM, MCI, and NC, were instructed to perform four proposed cognitive tasks whilst their EEG signals were collected. Statistical analysis was applied to determine the differences in oscillating patterns among the subject groups. EEG signals were then pre-processed, feature-extracted, and classified into their respective classes. Data acquisition procedures, statistical analysis, feature extraction techniques, classification method, and performance evaluation criteria are elaborated as follows.

\subsection{Participants}
A total of 45 volunteered participants, 15 per each subject group, were recruited from the memory clinic at Siriraj Hospital, Thailand. 
The types of dementia diagnosed in the DEM participants were varied, including early-onset dementia, Alzheimer’s Disease (AD), Dementia with Lewy bodies (DLB), Vascular Dementia (VaD), and Semantic Dementia (SD), with both mild and moderate severity stages.
The dementia stages were determined according to the International Classification of Diseases (ICD-10) of the WHO \cite{icd-10} and the diagnostic criteria of dementia from the Diagnostic and Statistical Manual of Psychiatric Disorders (DSM-V) \cite{dsm-v}.
\textcolor{black}{
The inclusion of various types of dementia was performed on a presumption that the common characteristics shared among dementia patients existed in both event related potential components and relative power based on knowledge from previous studies \cite{dkcokck2021, ltmsblyobtt2020}.
In order to develop a system to preliminary screen a large group of dementia, we would like to train the classification model with various types of dementia diagnosed participants.
}

The volunteered participants underwent thorough clinical neuroimaging and neurological examinations, including MRI, blood test, health check, neuropsychological test, and the Thai version of Mini-Mental-Status examination (TMSE), were conducted by licensed clinical psychologists to evaluate the cognitive functions. 
The clinical history and family background were also considered. Participants with the history of drug usage, brain injury, sickness, and severe dementia were excluded. 
The participants were required to wash and brush their hair and refrained from using any hair styling products on the visit. 
In addition, the consumption caffeine was prohibited prior to the EEG recording. 
All participants retained normal or corrected-to-normal visions. 
The characteristics of the participants are summarized in Table~\ref{tab:subj_characteristics}.

This study was performed with the approval of the Ethical Committee of the Faculty of Medicine, Siriraj Hospital, and Mahidol University, Thailand (COA No. SI 779/2019). Following the requirements of the Code of Ethical Principles for Medical Research Involving Human Subjects of the World Medical Association (Declaration of Helsinki), all participants or their legal representatives voluntarily partook and signed a written consent form with adequate information of the purpose and procedure prior to the study commencement. All the participants received monetary compensation.


\begin{table}[t]
\centering
\caption{Demographic Data of Subjects.}
\renewcommand{\arraystretch}{1.2}
\begin{tabular}{l l l l} \toprule[0.2em]
& NC (n=15)    & MCI (n=15)    & DEM (n=15)    \\ \midrule[0.1em]
Gender (M:F)             & 3:12         & 5:10          & 6:9           \\
Age (Years)              & 59.4 $\pm$ 9.4   & 70.7 $\pm$ 7.3    & 71.3 $\pm$ 9.0    \\
Education Level (Years)  & 16.3 $\pm$ 2.6   & 14.8 $\pm$ 3.4    & 11.1 $\pm$ 5.8    \\
Disease Duration (Years) & -            & 3.5 $\pm$ 3.4     & 4.5 $\pm$ 2.5     \\
TMSE scores              & 29.5 $\pm$ 0.8   & 27.3 $\pm$ 2.7    & 21.7 $\pm$ 7.0    \\ \bottomrule[0.2em]
\end{tabular}
\label{tab:subj_characteristics}
\end{table}

\subsection{Protocols and experiment design}

\subsubsection{Stimulus Presentation}
Visual stimuli were presented while the participants were seated at a distance of 60 cm from a 24" LG monitor (1920 x 1080 pixels, 60 Hz refresh rate). The size of the stimuli was set at 150 x 150 pixels, which accounted for approximately 3 degrees of the visual field. The presentation of the stimuli was designed and developed in Matlab using Psychtoolbox\footnote{http://psychtoolbox.org/}. The data were stored with the fast port I/O using io64 object\footnote{http://apps.usd.edu/coglab/psyc770/IO64.html}. The stimulating objects included basic symbols, such as X, plus, triangle, rectangle, pentagon, circle, square, and star. The on-screen instructions were written in simplified language, both in Thai and English, suitable for all educational levels.

\subsubsection{Experimental Procedure}
The participants performed four visually stimulated cognitive tasks: Fixation, Mental Imagery, Symbol Recognition, and Visually Evoked Related Potential (VERP).
\textcolor{black}{
In order to reduce performance bias, the sequence of the tasks being performed were managed in a pseudo-random order, in which the tasks were equally placed in the sequence.
}
The tasks were thoroughly inspected and reviewed by experts to ensure the suitability for cognitive assessment of the neurodegenerative patients. At the start of each task, the research staff explained the task, checked the participant's comprehension of the task, and continuously monitored the participant throughout the experiment.

\begin{figure}
     \centering
     \subfloat[Screen Positions \label{fig:screen_position}]{\includegraphics[width=0.3\linewidth]{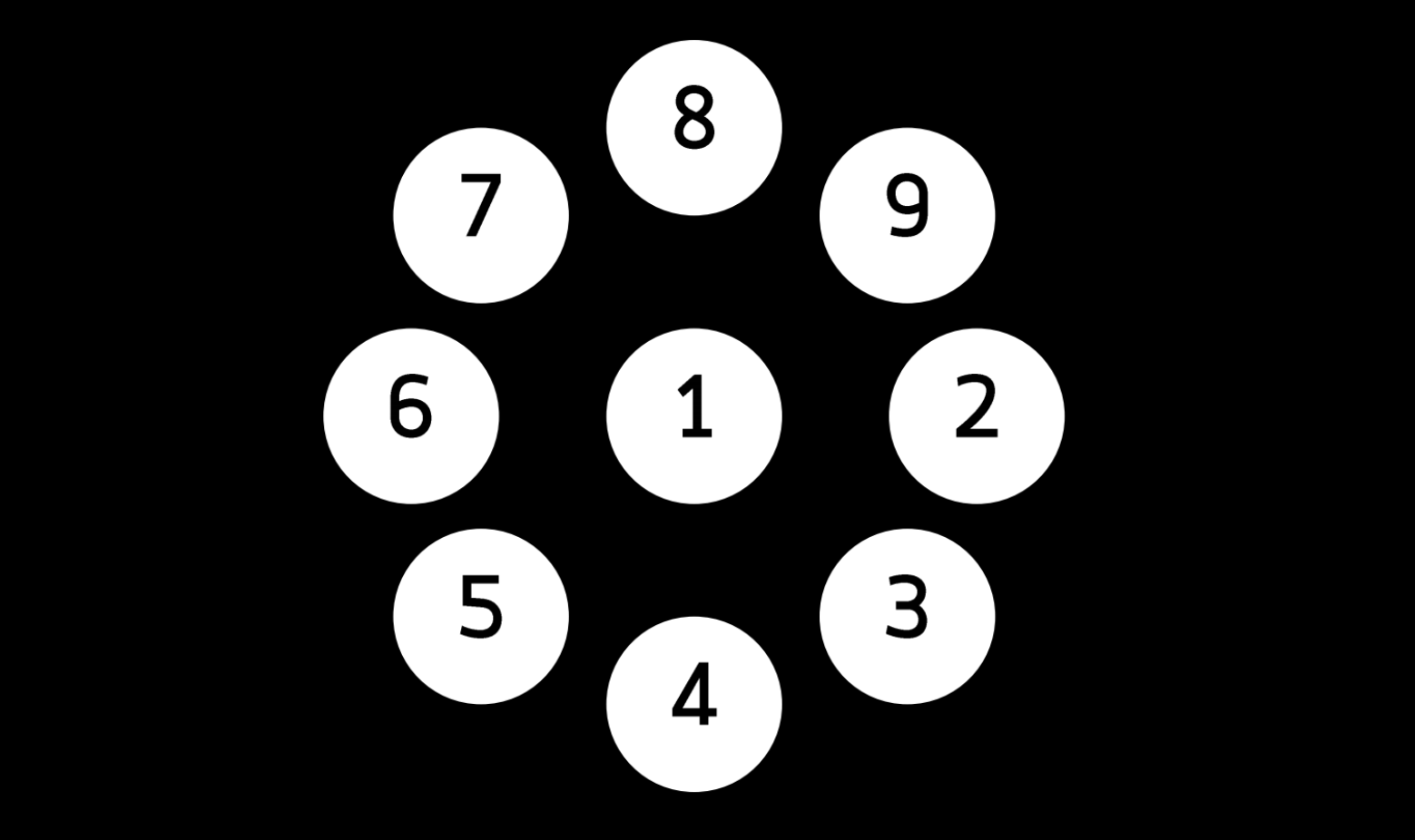}}
     
     \subfloat[Fixation Task \label{fig:fixation_task}]{\includegraphics[width=0.7\linewidth]{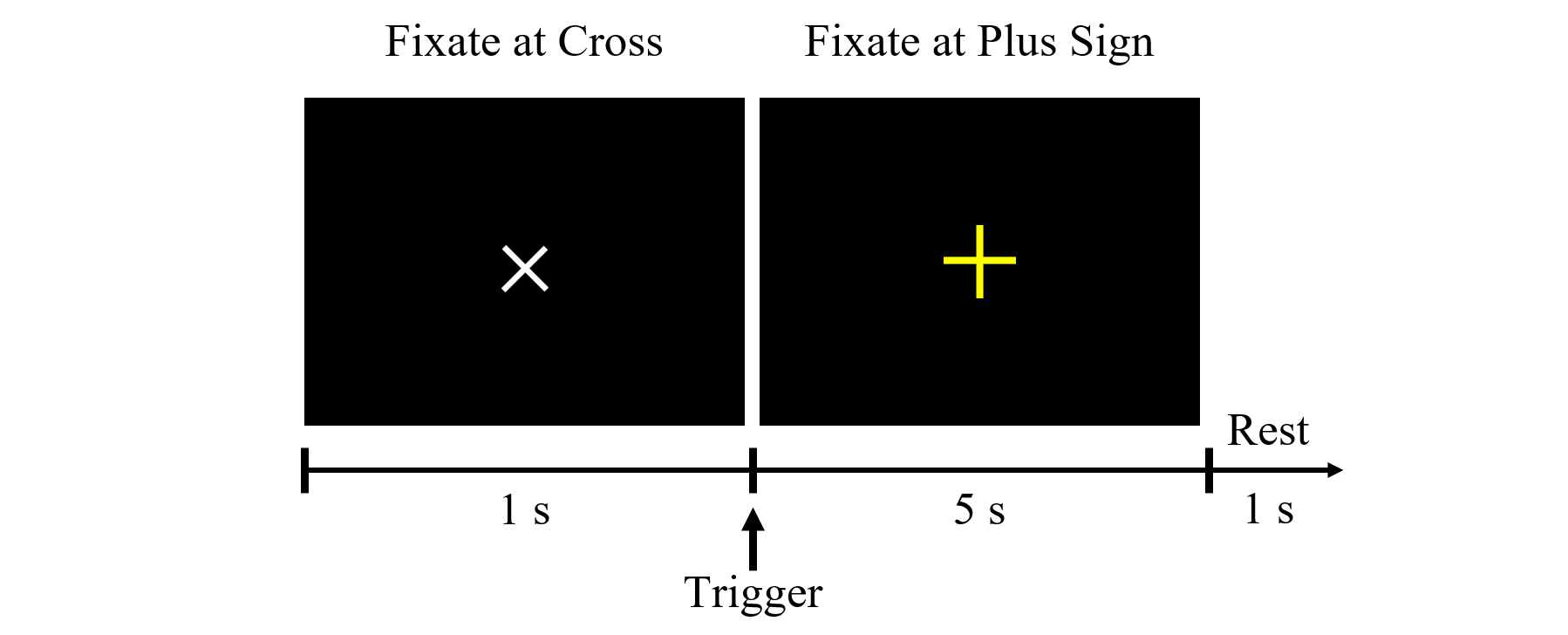}}
     
     \subfloat[Mental Imagery Task \label{fig:mi_task}]{\includegraphics[width=0.7\linewidth]{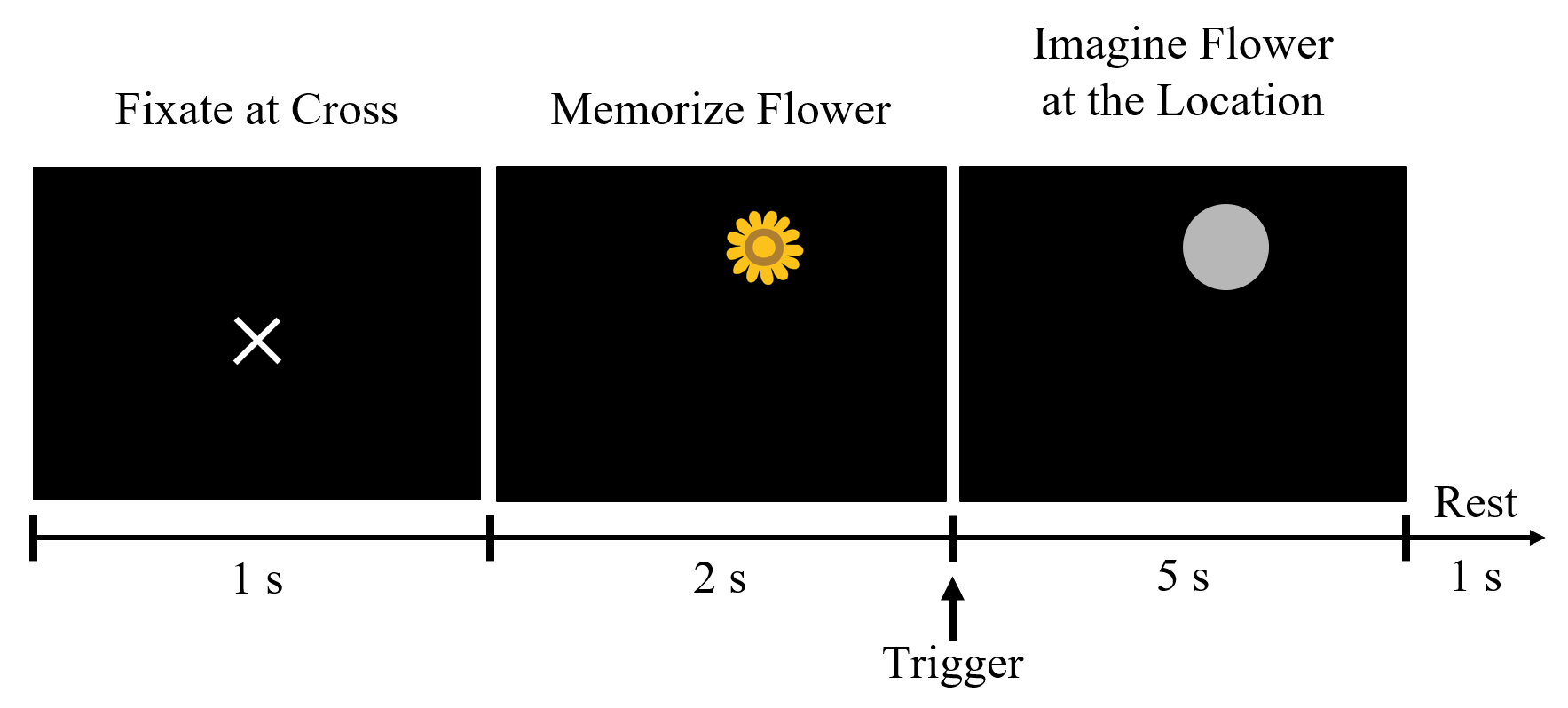}}
     
     \subfloat[Symbol Recognition Task \label{fig:symbol_task}]{\includegraphics[width=0.7\linewidth]{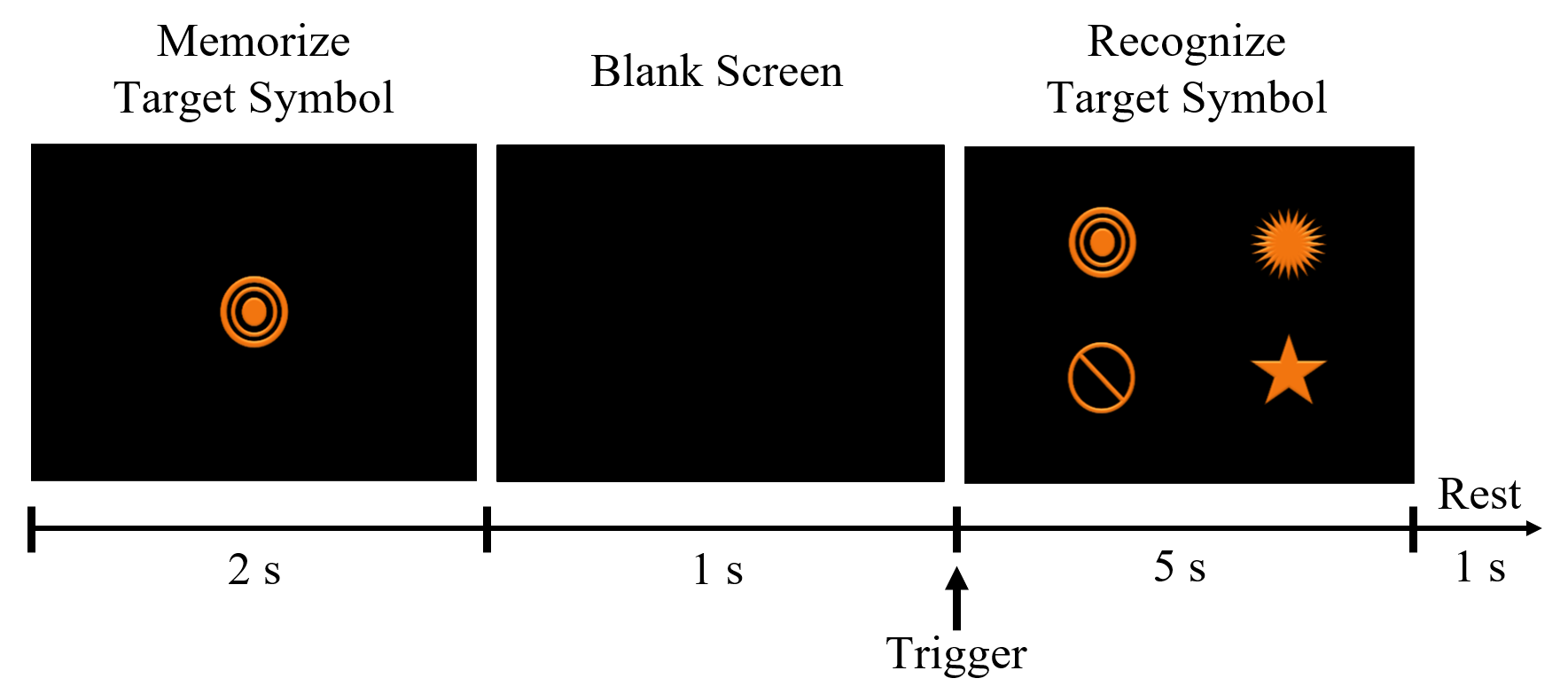}}
     
     \subfloat[Visually Evoked Related Potential Task \label{fig:verp_task}]{\includegraphics[width=0.7\linewidth]{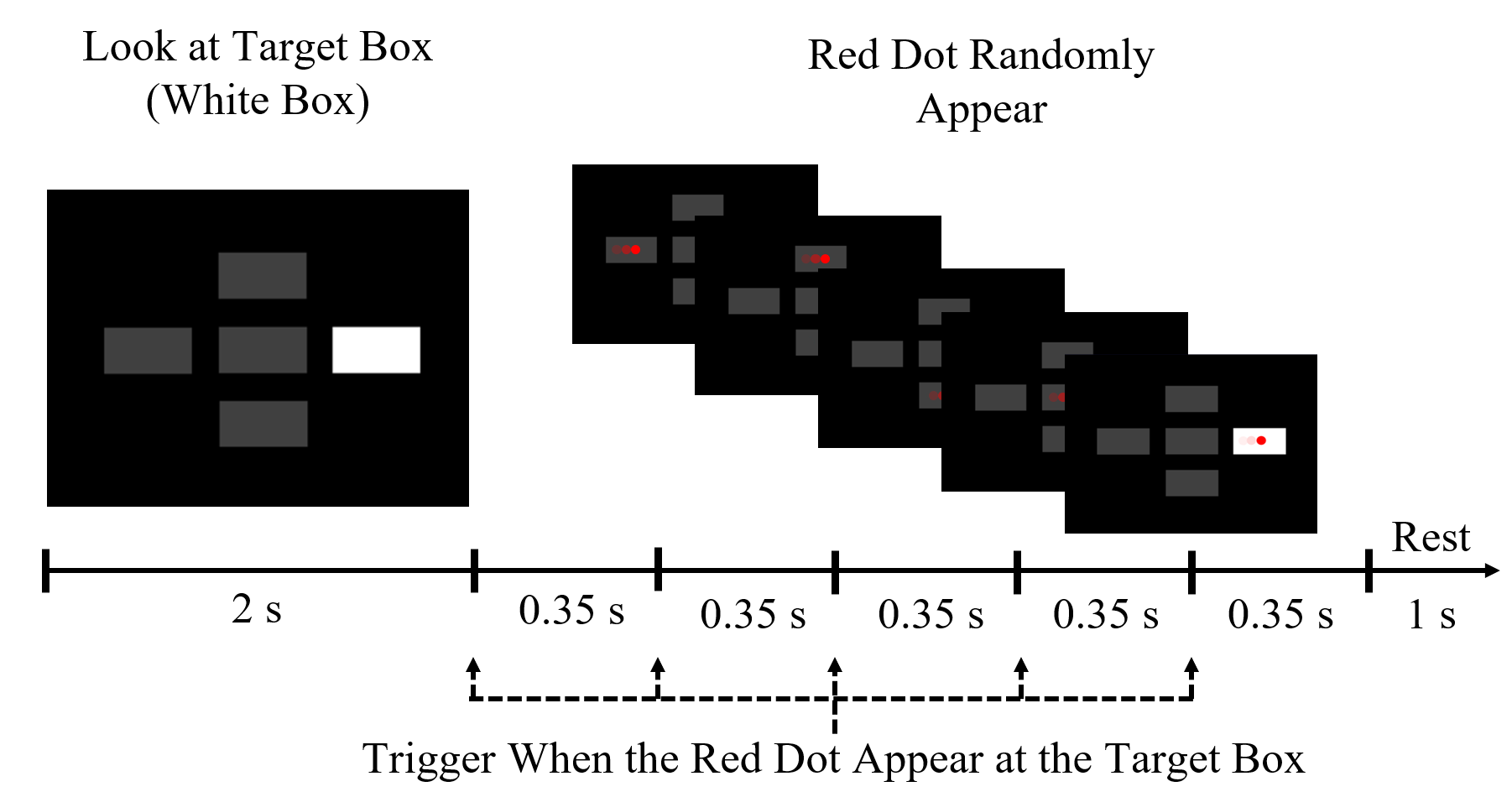}}
     
    \caption{Example of cognitive tasks and screen positions.}
    \label{fig:cognitive_tasks_example}
\end{figure}

\begin{table*}[ht!]
\centering
\caption{Cognitive Task Descriptions.}
\renewcommand{\arraystretch}{1.2}
\begin{tabular}{l p{0.2\linewidth}  p{0.35\linewidth} l} \toprule[0.2em]

Task & Neurocognitive Domains & Assumption & Ref    \\ \midrule[0.1em]
Fixation & Attention & 1. Increase of power in Theta band in MCI and DEM & \cite{ferbkyw2020} \\
& & 2. N2/P3 complex delayed in MCI and dementia & \cite{bmadf2016}   \\
Mental Imagery & Working Memory  & 1. Difference of power in Alpha band, mainly at parietal regions       & \cite{kll2018}  \\
& & 2. N170 delayed and reduced in MCI and dementia  & \cite{ggsh2008, bmadf2016}  \\
Symbol Recognition & Working Memory & 1. Increase of power in Theta and Alpha band in DEM & \cite{tkn2017, hwjw2017}\\
& & 2. Different P3 amplitude and latency   & \cite{hscssk2018, morrison2019visual}  \\
Visually Evoked Related Potential   & Executive Function (Decision Making)  & 1. Difference of ERPs responses within range of N200 and P200 components, mainly from temporo-occipital and parietal regions & \cite{kuba2007motion, ghgg2008,hscssk2018} \\ \bottomrule[0.2em]

\end{tabular}
\label{tab:tasks}
\end{table*}

The presentation screen was divided into nine positions, one at the center, and the remaining eight were evenly spaced surrounding the center, as shown in Figure~\ref{fig:screen_position}. The participants performed 30 trials in total for each task. The first three trials of each task served as testing trials to familiarize the participants with the instructions and the tasks, leaving 27 trials for analysis. The summary of cognitive domains and assumptions of four tasks are shown in \autoref{tab:tasks}.

\textbf{Fixation Task:} 
A fixation white cross was set to appear in the center of the screen with a black background prior to all the trials. In each trial, a yellow cross appeared at one of the nine positions on the screen. The sequence of appearing held a pseudo-random order with the same probability at all positions, as shown in Figure~\ref{fig:fixation_task}. The participants were instructed to fixate their gaze, while resisting blinking, for five seconds on the appearing the targeted yellow cross as soon as the cross was noticed. After each trial, the screen was cleared, and the next trial began with the yellow cross reappeared in a different position. While performing the task, the participants were encouraged to relax and refrain from engaging in any specific thoughts. As the task, focusing on capturing attention, did not require high cognitive load, the obtained EEG signals were treated as eye-opened resting-state EEG recording for baseline.

\textbf{Mental Imagery Task:} 
Inspired by a study from Kosmyna \textit{et al.} \cite{kll2018}, the Mental Imagery task was used to examine the working memory of the participants. Each trial began with an image of a simple flower on a black background. The position of the image was set to be pseudo-randomly located on one of the surrounding the nine positions. The participants were given two seconds to memorize the image, before being replaced by a gray circle. The gray circle remained for five seconds, providing a time window for the participants to mentally visualize the image at the location of the gray circle. All symbols disappeared at the end of each trial. An example of the Mental Imagery task is displayed in Figure~\ref{fig:mi_task}.

\textbf{Symbol Recognition Task:}
Inspired by a study from Han \textit{et al.} \cite{hwjw2017}, the Symbol Recognition task examined the working memory (for object recognition) and attention. The participants were instructed to memorize a target symbol presented for two seconds on a black background. A set of four random symbols with similar color and size, including the original target symbol, were presented on the screen after a one-second delay, as shown in Figure~\ref{fig:symbol_task}. The participants were given five seconds to select the target symbol with their gaze. Different target and non-target symbols appeared pseudo-randomly in each trial.

\textbf{Visually Evoked Related Potential Task (VERP):}
VERP task is known for the inspection of attention, decision making, and executive function, such as inhibitory control \cite{ghgg2008}. Four gray rectangular boxes (non-target boxes) and one white box (target box) were presented on a black background when each trial began. These five boxes were located on positions 1, 2, 4, 6, and 8 on the screen. The participants were instructed to concentrate on the target box for one second. A red dot then appeared inside one of the boxes, moved in a linear path from left to right, and disappeared when it reached the edge of the box. The participants were instructed to respond as fast as possible only to the red dot appearing inside the target box by following its movement with their gaze until it vanished. Conversely, the red dot that appeared on any of the non-target boxes was to be ignored. In each trial, the moving red dot appeared in all boxes, once per box, in a pseudo-random order. The position of the target box was altered randomly while maintaining the same probability in all positions. An example of VERP task is displayed in Figure~\ref{fig:verp_task}.

\begin{figure*}[!ht]
    \centering
    \includegraphics[width=1.4 \columnwidth]{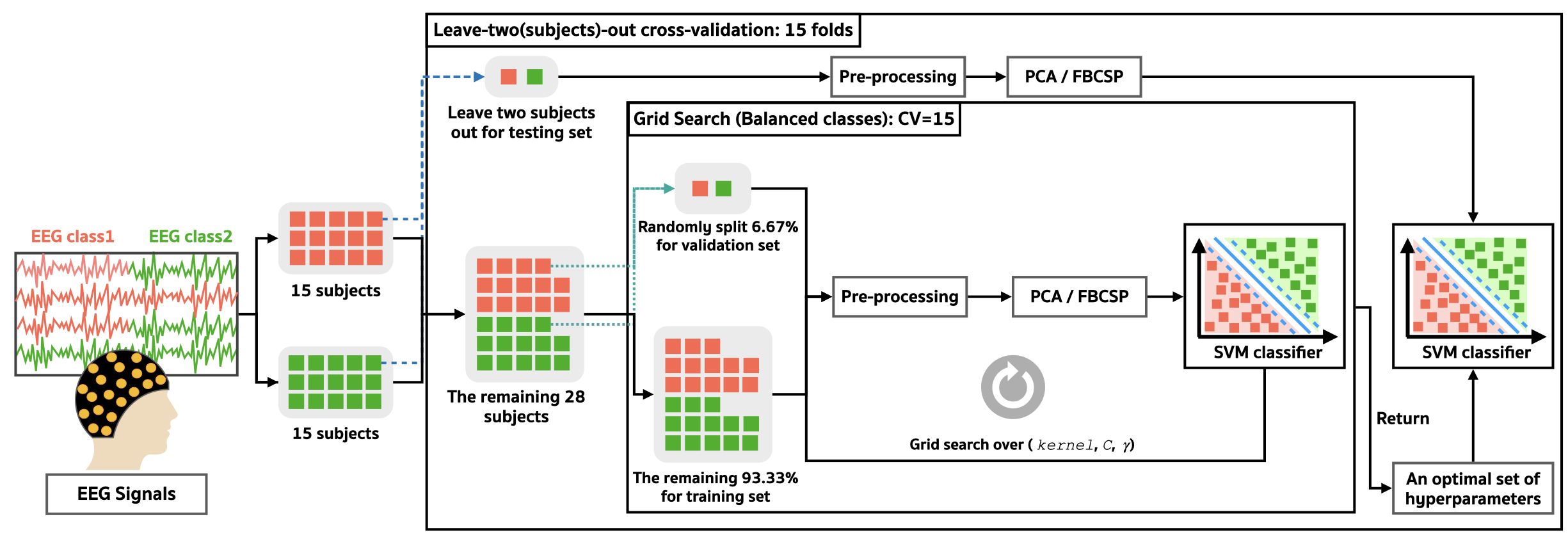}
    \caption{Framework of leave-two (subjects)-out cross-validation (LTOCV) with the grid search algorithm for the binary classification models.}
    \label{fig:fbcsp}
\end{figure*}

\subsection{\textcolor{black}{EEG Recordings and Pre-processing}}
The EEG data were recorded using a 32-channel active electrode system (two 16-channel) with two g.USBamp Research EEG amplifiers (g.tec medical engineering GmbH Austria), sampled at 256 Hz.
Thirty-two active electrodes were equidistantly placed over the scalp to cover all brain regions according to the 10-20 system, with a ground electrode at FPz and reference to the right earlobe. Conductivity gel was used to reduce noise and increase electrode conductivity. During the recording, the impedance of all electrodes was kept below 5 k$\Omega$. 
\textcolor{black}{Throughout the period, research staff continuously observed brain signal in a separate computer monitor, along with facial and physical gestures of the participants.
When any sign of drowsiness (for example, abnormal slow moving EEG signals, hands, legs, or body movements to prevent drowsiness) seemed to happen, warnings were given verbally to the participants to prevent drowsiness during EEG acquisition.
After completing each task (five minutes at most), the staff frequently checked the alertness of the participant and asked him/her to be relaxed or whether a short 1-minute pause was needed.}

The EEG signals recorded from the proposed tasks were pre-processed and analyzed using MATLAB, and EEGLAB toolbox \cite{dm2004}. The Notch filter at 50 Hz and band-pass filter at 1 to 30 Hz with a zero-phase shift FIR filter were applied to the raw data. Channels with unusable data periods were removed based on manual inspection. The removed channels were interpolated using nearby electrodes. The artifacts in the signal, including eyes and muscles, were rejected using the independent component analysis (ICA) \cite{dimigen2020optimizing} algorithm. Each signal was extracted into 30 epochs with 500 ms before stimulus onset to the end of the trial period, as described in \autoref{fig:cognitive_tasks_example}, with baseline correction. The first three epochs extracted from three testing trials were removed, leaving 27 epochs for analysis. The length of an epoch varied depending on the task. The features were then extracted from 30 EEG channels. The signals from Fp1 and Fp2 channels were excluded as they contained a higher amount of eye artifacts.

The same set of data was pre-processed with adjusted parameters deemed suitable for the event-related potentials (ERPs) analysis. The EEG signals were cleaned with the band-pass filtering at 1 to 30 Hz \cite{luck2014introduction}. The EEG epochs were extracted from 200 ms before to 800 ms after the onset of the stimulus. We removed the baseline using data from 200 ms before the onset of the stimulus.
\textcolor{black}{Then, for each task, every three consecutive trials were averaged to reduce the unexpected noise. Therefore, nine samples per subject were acquired for feeding into the machine learning model.}

\subsection{Statistical Analysis}

\subsubsection{Event Related Potentials (ERPs)}
After the obtained EEG traces were being segmented and pre-processed, the Kruskal-Wallis \cite{kw1952} test was performed to evaluate the significant difference of the ERPs signals between the three groups with a significance level of $p<0.01$. From the statistical test, the continuous-time windows of at least 31 ms long were determined as significant intervals and highlighted in gray, as shown in \autoref{fig:gaverage_erp}.

\subsubsection{Relative Power} \label{section:frequency_domain_stat}
The obtained EEG signals were filtered into four frequency bands of interest: delta (1-4 Hz), theta (4-8 Hz), alpha (8-13 Hz), and beta (13-30 Hz). Due to the muscle activity and artifacts, the gamma oscillation was excluded from our analysis \cite{mehjmnoswwa2018, wlpfcldwbw2008}. 
\textcolor{black}{
To determine the difference of powers in different oscillations between the three subject groups, the relative powers \cite{mbsmbbm2017,blwlyl2014} of each frequency band were calculated from the power spectral density (PSD) by the following equation:
\begin{equation}
    P_{relative}(f_1, f_2) = \frac {\sum_{f=f_1}^{f=f_2} P(f)}{\sum_{f=f_L}^{f=f_H} P(f)}
\end{equation}
where $ P(f) $ is the power at the specified frequency, $ f_1, f_2 $ indicate low and high frequencies of the interested frequency band, and $ f_L, f_H $ indicate 1 and 30 Hz, respectively.
}

The distributions of relative powers in the four frequency bands are presented in \autoref{fig:tasks_stat_frequency_domain}. 
The marked asterisk (*) among the pairs referred to the significant difference, tested with the Wilcoxon Rank Sum Test \cite{Haynes2013} at a significance level of $p<0.01$.


\subsection{Feature Extraction and Classification} 

\subsubsection{ERPs-PCA-SVM} \label{subsubsection:pca}
\textcolor{black}{
After being pre-processed, the total dimension of the data for ERPs analysis was represented by $n\_subjects \times n\_avg\_trials \times n\_channels \times n\_sampled Time Points$ ($45\times9\times30\times256$) for feeding into machine learning model.
For classification, the signals were selected from the stimulus onset to 800 ms after, resulting in 205-time points.
In total, there are 6,150 temporal features (205-time points $\times$ 30 EEG channels) for each task. 
Principal Components Analysis (PCA) has been widely accepted as a dimension reduction method for EEG in both temporal and frequency domain \cite{barry2018eeg}. 
In this work, PCA was applied to decompose the meaningful components from 6,150 features with more than 90\% of the variance. 
Those principal components were used as the inputs of the SVM for the classification of the subjects.
Overall of classification framework is illustrated in \autoref{fig:fbcsp}.}

\subsubsection{Relative Power-FBCSP-SVM}
After being pre-processed, there are 27 trials remained per subject.
For Fixation, Mental Imagery, and Symbol Recognition tasks, 1,280 time points (256 Hz $\times$ 5 seconds) were selected from the onset of the stimulus and ended at 5 seconds after, \textcolor{black}{resulting in a dimension of $n\_subjects \times n\_trials \times n\_channels \times n\_sampled Time Points$ ($45\times27\times30\times1,280$).}
For VERP task, 448 time points (256 $\times$ 1.75 seconds) were selected from the onset and ended at 1.75 seconds (or 105 ms) after, \textcolor{black}{resulting in $45\times27\times30\times448$.}

Filter Bank Common Spatial Pattern (FBCSP) was performed to extract the features from the EEG signals for cognitively impaired patient classification.
This algorithm is based on the Common Spatial Pattern (CSP) algorithm and was initially employed as a feature extraction method for classifying 2-class motor imagery EEG data \cite{aczg2008, camlktmmw2020}.
We examined the algorithm’s ability to successfully extract features from EEG data for binary class cognitively impaired patient classification.
\textcolor{black}{
With the algorithm's ability to capture frequency features from EEG, we had employed this as a feature extraction algorithm.
}
Additionally, we constructed five SVM models using this algorithm. 
Each model of the first four models used the features from each cognitive task, while the fifth model used features from all four cognitive tasks to determine the most suitable task to detect those cognitively impaired.

For this approach, FBCSP algorithm was implemented using MNE-Python package \cite{mne2013}.
Four spatial filters were applied to decompose EEG signals into four frequency bands.
\textcolor{black}{
Closely examined from the distributions of relative power, for each task and each classification pair, those statistically significant frequency bands were being selected and fed to the algorithm.}
Later, from those determined bands, the algorithm selected the discriminative EEG features based on a set of pre-specified indicators.
The algorithm comprised of four stages: band-pass filtering, spatial filtering with the CSP algorithm, selection of features, and classification \cite{acwgz2012}.
The selected features were used as inputs of an SVM to classify the signals into their respective classes.
The overall process is illustrated in \autoref{fig:fbcsp}.

\subsection{Performance Evaluation Method} \label{subsection:performance_evaluation_method}

To evaluate the classification performance, we used Leave-two (subjects)-out cross-validation (LTOCV) for 15 folds.
\textcolor{black}{
Initially, the data from \textit{N} subjects of binary classes were separated into a set of \textit{N-2} subjects for training and validating (a ratio split training:validating of 9:5), and a set of 2 subjects (1 from each class) for testing.}
This cross-validation mechanism was conducted with Python Scikit-learn library \cite{sklearn_paper}.
We performed hyperparameter tuning for the SVM, using the Grid Search Algorithm \cite{gridsearch_paper} with the validation set. The hyperparameter space of the SVM in this study consisted of the set \{radial basis function (RBF), sigmoid, linear\} for kernel; the set $\{ 0.001, 0.01, 0.1, 1, 10, 100, 1000\}$  for $C$; and the set  $\{0.001, 0.0001, 0.00001, 0.000001\}$ for gamma. The process was repeated until every subject was selected for testing. As for the metrics, we used the accuracy (or correct rate), sensitivity (True Positive Rate), and specificity (True Negative Rate), providing an insight on the accuracy and robustness of the classifier. The evaluation scores ranged from 0 to 1, where 0 was the worst and 1 was the perfect classification. The resulting evaluation scores were computed by taking the average results of all subjects. For each subject, we computed the performance score by averaging classification results over the trials.

\section{Results}
\label{sec_results}
Statistical tests revealed the existence of differences among the groups. Machine learning-based algorithms were employed to test the efficiency of two techniques: ERPs-PCA-SVM and Relative Power-FBCSP-SVM. Among these techniques, extracting features with Relative Power and FBCSP produced better performance. To investigate the effectiveness of Relative Power-FBCSP-SVM, we further explored the suitable number of trials that would yield an acceptable accuracy by varying the minimum number of trials required for the cognitive task. The results led to the acquisition of high-quality cortical signals from sufficient trials using this proposed framework with higher operability and lesser burden on the patients.

\begin{figure*}[t]
     \centering
     \subfloat[Fixation]
     {\includegraphics[width=0.25\linewidth]{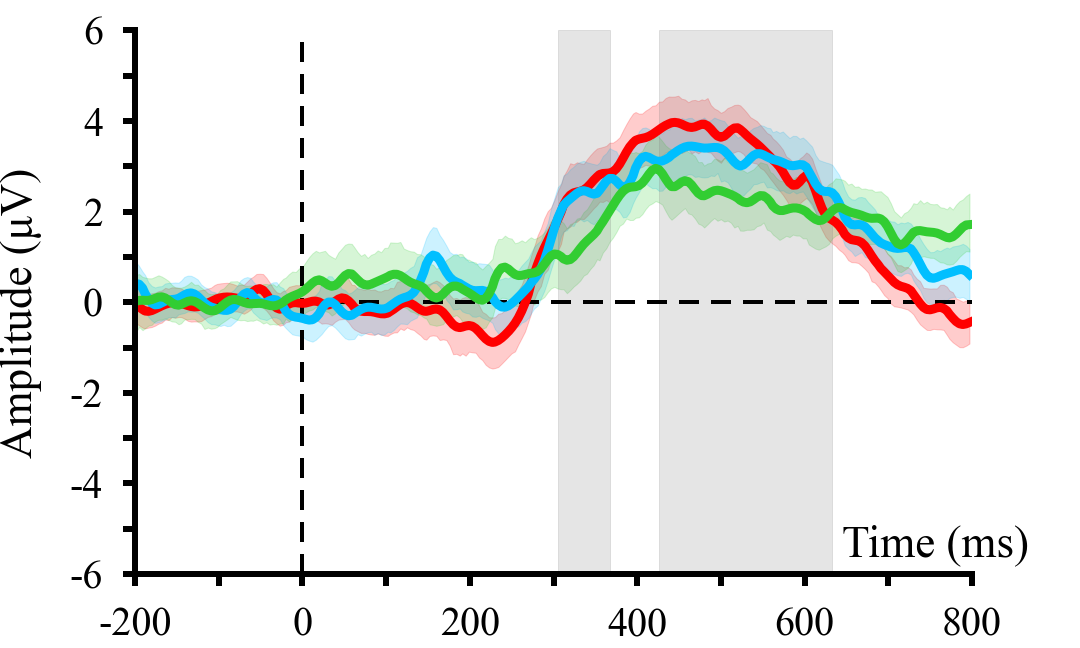}\label{subfig:task1_Pz}}
     \subfloat[Mental Imagery]
     {\includegraphics[width=0.25\linewidth]{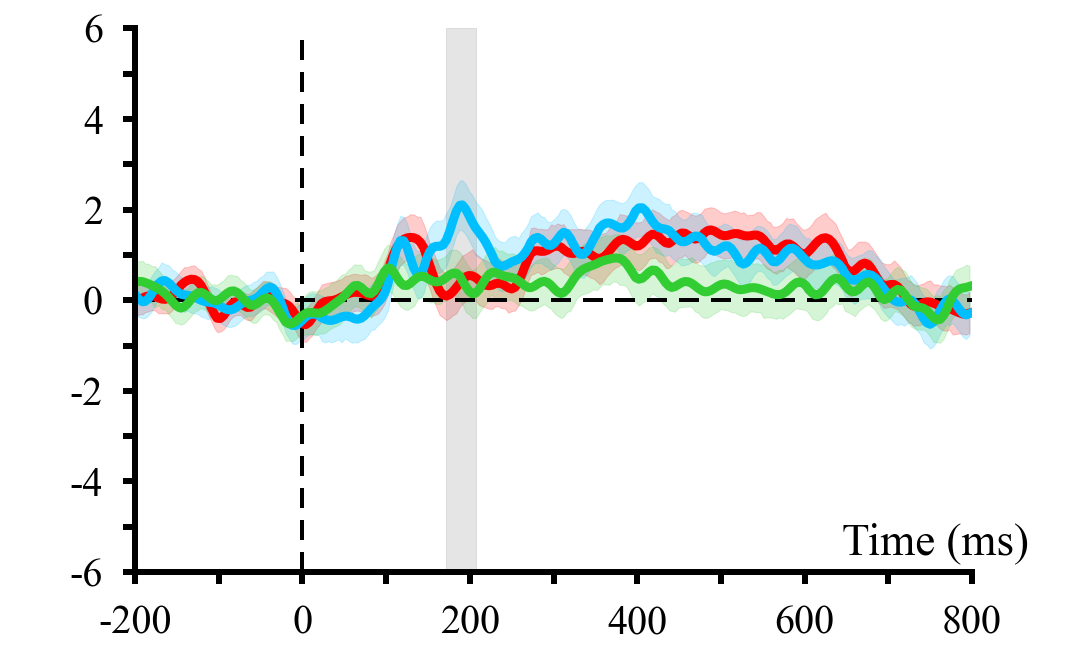}\label{subfig:task10_Pz}}
     \subfloat[Symbol Recognition]
     {\includegraphics[width=0.25\linewidth]{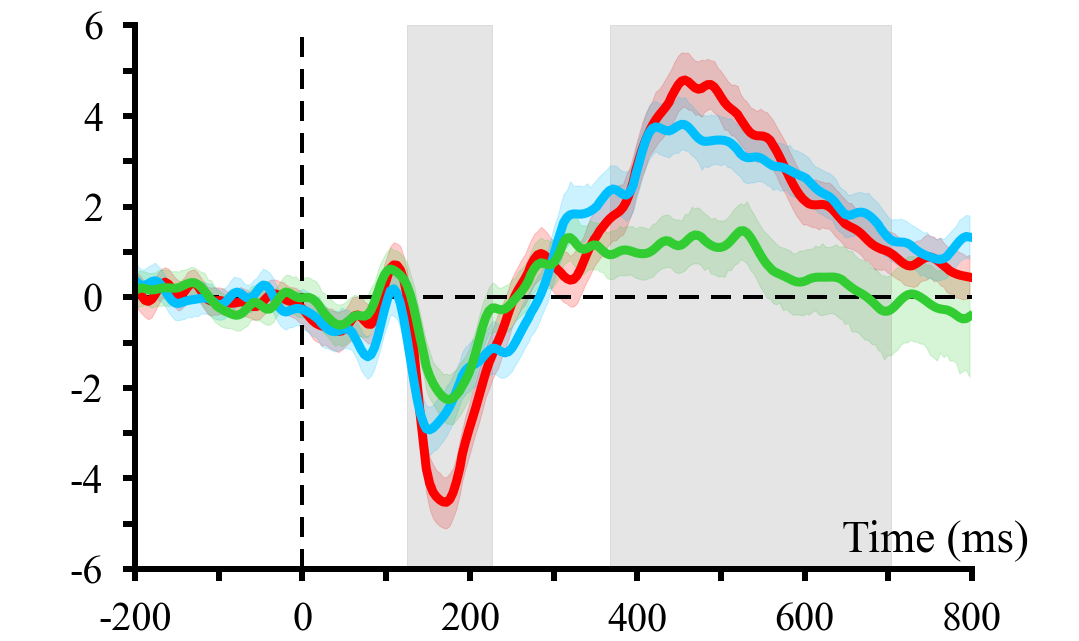}\label{subfig:task11_Pz}}
     \subfloat[VERP]
     {\includegraphics[width=0.25\linewidth]{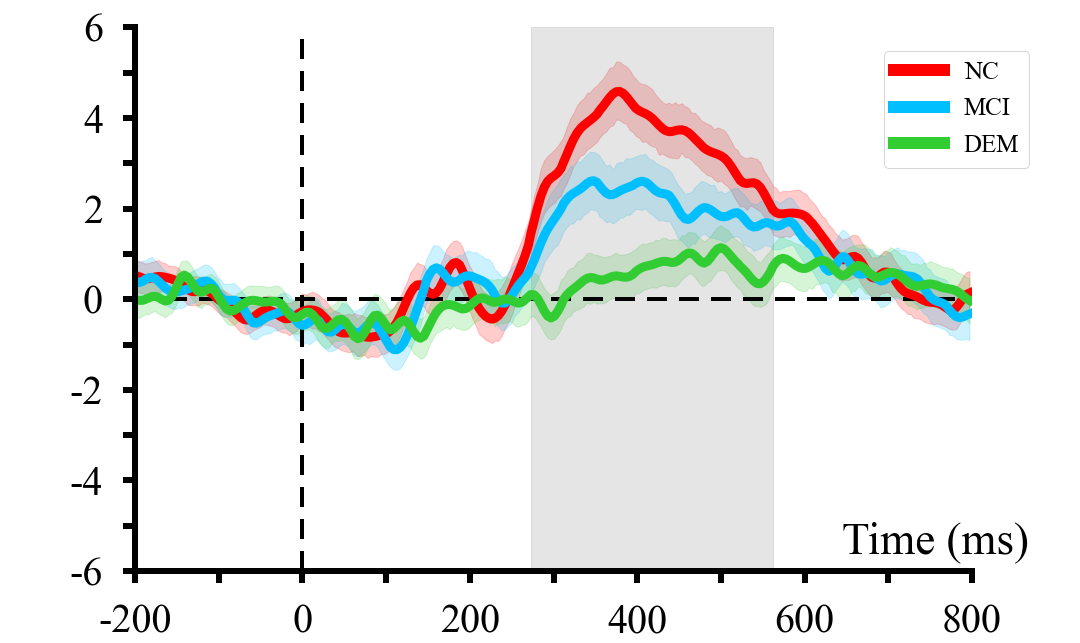}\label{subfig:task12_Pz}}
     
    \caption{The grand average of the time-locked responses recorded at Pz location from different subject groups during (a) Fixation Task, (b) Mental Imagery Task, (c) Symbol Recognition Task, and (d) Visually Evoked Related Potential Task (VERP). \textcolor{black}{The color bars indicate the confidence interval of 95\% }while the additional gray bars show the significant difference of at least two groups corresponding to the Kruskal Wallis test with the specific criteria ($p<0.01$).}
    \label{fig:gaverage_erp}
\end{figure*}

\subsection{Statistical Results}
\subsubsection{Event Related Potentials (ERPs)}
The significant intervals in the ERPs plot were observed in the EEG signals from 30 channels. 
\textcolor{black}{
For each channel and task, grand average of EEG responses among subjects in the group were computed.}
The intervals of the average of EEG responses among the three subject groups differed significantly, according to the Kruskal-Wallis test. An example of the statistical results at the Pz electrode is shown in \autoref{fig:gaverage_erp}, in which the significant intervals are highlighted in gray, \textcolor{black}{and the color bars indicate 95\% confidence interval.}

In the Fixation task, the significant intervals that contained negative peaks were observed in all subject groups at the electrodes positioned in the parietal lobe at P7, P9, POz, P6, and P8 (see Supplementary Figure S1). The positive peaks found in long significant intervals of more than 200 ms at the CP3, CPz, and Pz channels corresponded to the late response P300. In these intervals, the declined amplitude of the ERPs signals was found only from DEM group.

No noticeable responses were found in the Mental Imagery task with the exception of the late positive peak, at the Pz, POz, and Oz locations, emerging from the MCI group (see Supplementary Figure S2). In addition, the negative peaks inside the short significant intervals at the P7 and P9 electrodes were found to be evoked. The amplitude of these negative peaks was the lowest in DEM group compared to the other groups.

For the Symbol Recognition task (see Supplementary Figure S3), the short and medium-length significant intervals with strong negative peaks elicited mainly at the parietal and occipital lobes in the range between 78 – 250 ms. The most prominent negative peaks were observed in the NC group. Additionally, the longer significant intervals with strong positive peaks in the range of 400 – 600 ms were observed at the CPz, CP4, Pz, and POz electrodes. The negative peaks with a large amplitude, reflecting mirror images to the previous positive peaks, were found to be evoked on the frontal lobe, especially at the F7, F8, F9, and F10 channels.

The ERPs signals elicited during the VERP task from each group were visually distinguishable (see Supplementary Figure S4). For all scalp electrodes, the amplitude of the ERPs signals from NC group appeared to be the largest, followed by MCI and DEM groups. In the range between 285 – 500 ms, the negative and positive peaks at the long significant intervals were observed mainly at the left frontal, midline, and right parietal lobes.  The significant intervals that contained N200 latency were found in a few channels on the parietal lobe (CP3, P6, and P8).

\subsubsection{Relative Power}
\begin{figure*}[ht!]
     \centering
    \includegraphics[width=2.1 \columnwidth]{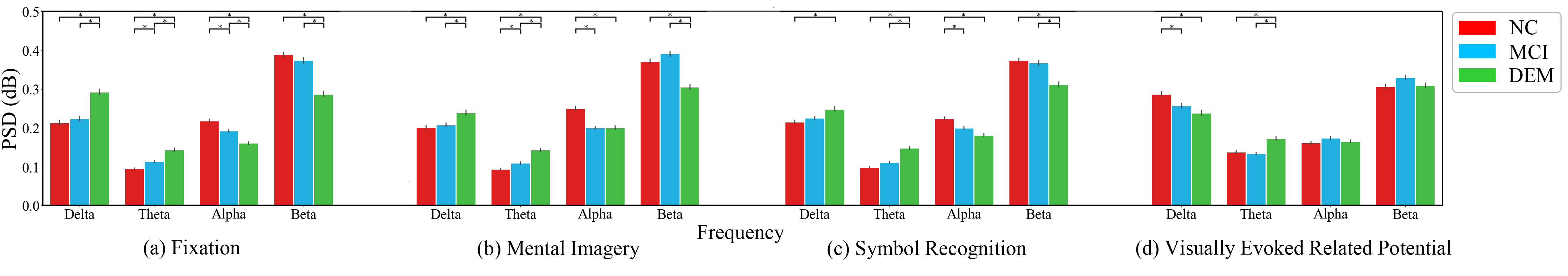}
    \caption{Distributions of power among the three subject groups during (a) Fixation Task, (b) Mental Imagery Task, (c) Symbol Recognition Task, and (d) Visually Evoked Related Potential Task (VERP) for all cortical regions in following frequency bands: Delta (1-4 Hz), Theta (4-8 Hz), Alpha (8-13 Hz), and Beta (13-30 Hz). * denotes the significance between a group-pair tested with the Wilcoxon Rank Sum Test at $p<0.01$.}
    \label{fig:tasks_stat_frequency_domain}
\end{figure*}

\begin{figure*}[ht!]
    \centering
    \includegraphics[width=1.3 \columnwidth]{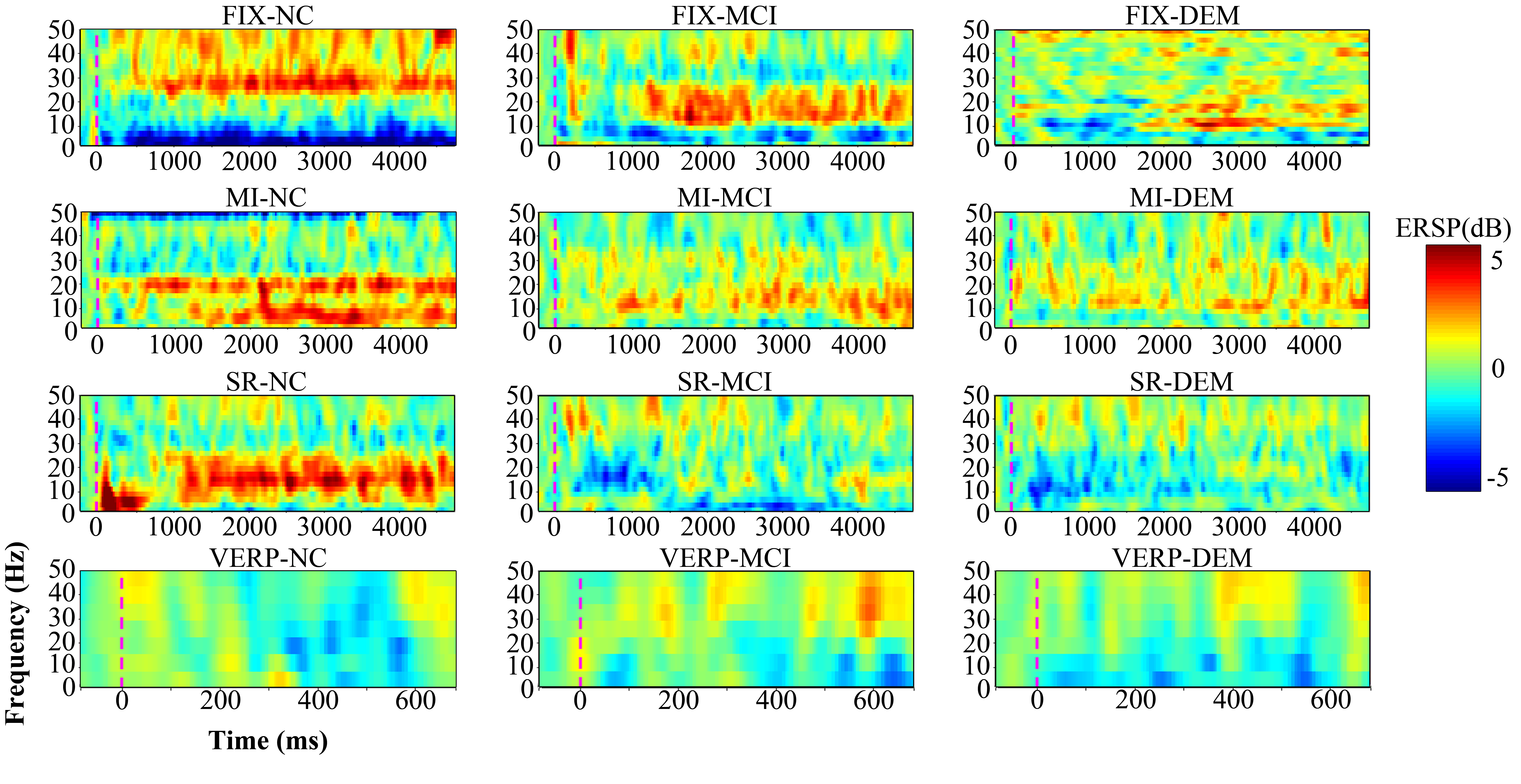}
    \caption{Analysis of EEG power in terms of Time-Frequency Distribution at Pz location measured by ERSP for different subject groups during the Fixation Task (FIX), Mental Imagery Task (MI), Symbol Recognition Task (SR), and Visually Evoked Related Potential Task (VERP).}
    \label{fig:ersp_figure}
\end{figure*}

The statistically significant differences between the EEG power in each band at the level of $p<0.01$, as mentioned in Section \ref{section:frequency_domain_stat}, are shown in \autoref{tab:stat_analysis_results} for each pair of the subject groups and each cognitive task.
The EEG powers of the three groups were statistically different in the theta and alpha bands for the Fixation and Metal Imagery tasks, whereas only the alpha band was found during the Symbol Recognition task.
No statistical significance was found in the bands among the three subject groups during the VERP task.
\textcolor{black}{Additionally, as shown in Figure \ref{fig:tasks_stat_frequency_domain}, in the first three tasks, the power in the slow-moving frequency bands (delta and theta) was found to be increased as cognitive declines.
On the other hand, the power in the fast-moving frequency bands (alpha and beta) was found to be in the opposite.}
Furthermore, during the VERP task, the EEG power of DEM group, in comparison to the NC group, increased in the theta band, but vice versa in the delta band.
The distributions of power in the time-frequency domain measured by Event-Related Spectral Perturbation (ERSP) \cite{makeig1993} for all tasks and groups are shown in \autoref{fig:ersp_figure}.

\begin{table}[]
\centering
\caption{Statistically Significant ($p<0.01$) Frequency Bands for Pair  Classification. $\delta$: Delta, $\theta$: Theta, $\alpha$: Alpha, $\beta$: Beta band. }
\renewcommand{\arraystretch}{1.2}
\begin{tabular}{l c c c c} \toprule[0.2em]
\textbf{Pair} & \textbf{FIX} & \textbf{MI} & \textbf{SR} & \textbf{VERP}\\ \midrule[0.1em]
NC-DEM & $\delta , \theta, \alpha, \beta $ & $\delta , \theta, \alpha, \beta $ & $\delta , \theta, \alpha, \beta $ &   $\delta, \theta $      \\
NC-MCI & $\delta , \theta, \alpha, \beta $  & $\theta , \alpha$    & $\alpha$ &  $\delta $ \\
MCI-DEM & $\theta, \alpha $  & $\theta , \alpha$ & $\theta, \beta $  & $\theta$  \\
\bottomrule[0.2em]
\end{tabular}
\label{tab:stat_analysis_results}
\end{table}

\subsection{Classification Results}

\subsubsection{ERPs-PCA-SVM}

\begin{table*}[t]
\centering
\caption{Classification Results (Accuracy $\pm$ SE, Sensitivity $\pm$ SE, and Specificity $\pm$ SE) from ERPs-PCA-SVM}
\renewcommand{\arraystretch}{1.2}
\resizebox{\textwidth}{!}
{\begin{tabular}{@{}lccccccccc@{}}
\toprule[0.2em]
\multicolumn{1}{c}{\multirow{2}{*}{\textbf{Task}}} & \multicolumn{3}{c}{\textbf{NC-MCI}}        & \multicolumn{3}{c}{\textbf{NC-DEM}}        & \multicolumn{3}{c}{\textbf{MCI-DEM}}       \\ \cmidrule[0.1em](l){2-4} \cmidrule[0.1em](l){5-7}\cmidrule[0.1em](l){8-10}  
\multicolumn{1}{c}{} &
  \multicolumn{1}{l}{\textbf{Accuracy}} &
  \multicolumn{1}{l}{\textbf{Sensitivity}} &
  \multicolumn{1}{l}{\textbf{Specificity}} &
  \multicolumn{1}{l}{\textbf{Accuracy}} &
  \multicolumn{1}{l}{\textbf{Sensitivity}} &
  \multicolumn{1}{l}{\textbf{Specificity}} &
  \multicolumn{1}{l}{\textbf{Accuracy}} &
  \multicolumn{1}{l}{\textbf{Sensitivity}} &
  \multicolumn{1}{l}{\textbf{Specificity}} \\ \midrule[0.1em]
Fixation              & 0.45 $\pm$ 0.04 & 0.41 $\pm$ 0.06   & 0.50 $\pm$ 0.06   & 0.64 $\pm$ 0.05 & 0.61 $\pm$ 0.06   & 0.67 $\pm$ 0.06   & 0.47 $\pm$ 0.05 & 0.47 $\pm$ 0.07   & 0.47 $\pm$ 0.08   \\
Mental Imagery        & 0.50 $\pm$ 0.04 & 0.45 $\pm$ 0.05   & 0.55 $\pm$ 0.05   & 0.59 $\pm$ 0.03 & 0.54 $\pm$ 0.06   & 0.64 $\pm$ 0.05   & 0.55 $\pm$ 0.04 & 0.50 $\pm$ 0.07   & 0.60 $\pm$ 0.07   \\
Symbol Recognition    & \textbf{0.61 $\pm$ 0.05} & \textbf{0.57 $\pm$ 0.06}   & 0.66 $\pm$ 0.07   & 0.76 $\pm$ 0.04 & 0.70 $\pm$ 0.09   & 0.81 $\pm$ 0.04   & \textbf{0.59 $\pm$ 0.05} & 0.53 $\pm$ 0.08   & \textbf{0.64 $\pm$ 0.06}   \\
VERP                  & 0.51 $\pm$ 0.06 & 0.46 $\pm$ 0.07   & 0.57 $\pm$ 0.08   & 0.72 $\pm$ 0.04 & \textbf{0.70 $\pm$ 0.04}   & 0.74 $\pm$ 0.06   & 0.50 $\pm$ 0.04 & \textbf{0.60 $\pm$ 0.06}   & 0.39 $\pm$ 0.08   \\
\hline \hline
4-Task Combination    & 0.59 $\pm$ 0.06 & 0.44 $\pm$ 0.08   & \textbf{0.74 $\pm$ 0.07}   & \textbf{0.77 $\pm$ 0.04} & 0.70 $\pm$ 0.06   & \textbf{0.84 $\pm$ 0.05}   & 0.56 $\pm$ 0.05 & 0.54 $\pm$ 0.08   & 0.58 $\pm$ 0.07   \\
\bottomrule[0.2em]
\end{tabular}}
\label{tab:erp_classification_result}
\end{table*}

\textcolor{black}{
We evaluated the performance of several SVM models for the classification of the three pairs (NC-MCI, NC-DEM, and MCI-DEM) with Leave-two (subjects)-out cross-validation (LTOCV). }
The results from the four individual tasks and the 4-task combination, along with the accuracy (acc), sensitivity (sen), and specificity (spec) with the standard errors (SE), are summarized in \autoref{tab:erp_classification_result}. 

According to those measurement values, SVM models performed best in classifying NC-DEM, using the 4-task combination features (acc: 0.77, sen: 0.70, and spec: 0.84). In contrast, the features from the 4-task combination appeared to produce lower measurement values than the features from the individual tasks detected in other pairs. Amongst the individual tasks, the performances during Symbol Recognition task in all three pairs yielded the best overall results. Additionally, a competitive capability for distinguishing between NC and DEM groups during the VERP task was revealed with accuracy, sensitivity, and specificity of 0.72, 0.70, and 0.74, respectively.

\subsubsection{Relative Power-FBCSP-SVM}

\textcolor{black}{
The SVM models were evaluated using LTOCV.}
The classification results of the three group pairs (NC-MCI, NC-DEM, and MCI-DEM) are reported in terms of different parameters (acc, sen, and spec) with SE for the four individual cognitive tasks and 4-task combination as well as the three classification pairs, as shown in \autoref{tab:fbcsp_classification_result}.

The task that provided the best classification of the NC-MCI pair was found to be VERP (acc: 0.69, sen: 0.77, spec: 0.60), whereas for the MCI-DEM pair was the Mental Imagery (acc: 0.72, sen: 0.66, spec: 0.77). Mental Imagery task (acc: 0.80, sen: 0.80, spec: 0.80) and the combination of all four tasks (acc: 0.81, sen: 0.87, spec: 0.75) produced comparable good results for classifying the NC-DEM pair. However, when monitoring the correctness and true positive rate, the combination of all four tasks yielded the best classification of the pairs.

\begin{table*}[t]
\centering
\caption{Classification Results (Accuracy $\pm$ SE, Sensitivity $\pm$ SE, and Specificity $\pm$ SE) from Relative Power-FBCSP-SVM}
\renewcommand{\arraystretch}{1.2}
\resizebox{\textwidth}{!}
{\begin{tabular}{@{}lccccccccc@{}}
\toprule[0.2em]
\multicolumn{1}{c}{\multirow{2}{*}{\textbf{Task}}} & \multicolumn{3}{c}{\textbf{NC-MCI}}        & \multicolumn{3}{c}{\textbf{NC-DEM}}        & \multicolumn{3}{c}{\textbf{MCI-DEM}}       \\ \cmidrule[0.1em](l){2-4} \cmidrule[0.1em](l){5-7}\cmidrule[0.1em](l){8-10}  
\multicolumn{1}{c}{} &
  \multicolumn{1}{l}{\textbf{Accuracy}} &
  \multicolumn{1}{l}{\textbf{Sensitivity}} &
  \multicolumn{1}{l}{\textbf{Specificity}} &
  \multicolumn{1}{l}{\textbf{Accuracy}} &
  \multicolumn{1}{l}{\textbf{Sensitivity}} &
  \multicolumn{1}{l}{\textbf{Specificity}} &
  \multicolumn{1}{l}{\textbf{Accuracy}} &
  \multicolumn{1}{l}{\textbf{Sensitivity}} &
  \multicolumn{1}{l}{\textbf{Specificity}} \\ \midrule[0.1em]
Fixation & 0.59 $\pm$ 0.05 & 0.63 $\pm$ 0.05 & 0.55 $\pm$ 0.05 & 0.62 $\pm$ 0.05 & 0.69 $\pm$ 0.05 & 0.54 $\pm$ 0.05 & 0.62 $\pm$ 0.03 & 0.56 $\pm$ 0.05 & 0.67 $\pm$ 0.04 \\
Mental Imagery & 0.55 $\pm$ 0.05 & 0.60 $\pm$ 0.05 & 0.50 $\pm$ 0.05 & 0.80 $\pm$ 0.03 & 0.80 $\pm$ 0.03 & \textbf{0.80 $\pm$ 0.03} & \textbf{0.72 $\pm$ 0.05} & \textbf{0.66 $\pm$ 0.05} & \textbf{0.77 $\pm$ 0.05} \\
Symbol Recognition & 0.49 $\pm$ 0.03 & 0.55 $\pm$ 0.05 & 0.42 $\pm$ 0.03 & 0.74 $\pm$ 0.05 & 0.79 $\pm$ 0.04 & 0.69 $\pm$ 0.06 & 0.63 $\pm$ 0.05 & 0.61 $\pm$ 0.05 & 0.65 $\pm$ 0.05 \\
VERP & \textbf{0.69 $\pm$ 0.05} & \textbf{0.77 $\pm$ 0.04} & \textbf{0.60 $\pm$ 0.05} & 0.65 $\pm$ 0.05 & 0.68 $\pm$ 0.04 & 0.61 $\pm$ 0.04 & 0.52 $\pm$ 0.06 & 0.53 $\pm$ 0.05 & 0.51 $\pm$ 0.05 \\
\hline \hline
4-Task Combination & 0.62 $\pm$ 0.04 & 0.67 $\pm$ 0.04 & 0.56 $\pm$ 0.05 & \textbf{0.81 $\pm$ 0.05} & \textbf{0.87 $\pm$ 0.05} & 0.75 $\pm$ 0.05 & 0.55 $\pm$ 0.05 & 0.53 $\pm$ 0.04 & 0.57 $\pm$ 0.06 \\
 \bottomrule[0.2em]
\end{tabular}}
\label{tab:fbcsp_classification_result}
\end{table*}

\subsubsection{Suitable Number of Trials}

Following the execution of Relative Power-FBCSP-SVM on all 27 cognitive task trials, we reached the speculations that performing a high number of trials during the cognitive tasks could cause mental fatigue and physical burden on the participants leading to poor classification results. We then proceeded with determining the most suitable number of trials that yielded the best accuracy.

\begin{figure*}[ht!]
     \centering
    \includegraphics[width=2.1 \columnwidth]{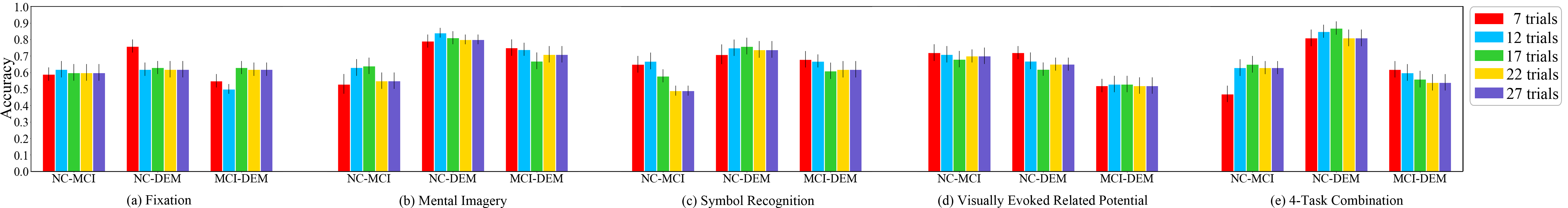}
    \caption{Classification performance evaluated in terms of accuracy for three classification pairs in (a) Fixation Task, (b) Mental Imagery Task, (c) Symbol Recognition Task, (d) Visually Evoked Related Potential Task (VERP), and (e) 4-task combination by varying number of trials required for Relative Power-FBCSP-SVM.}
    \label{fig:result_combo}
\end{figure*}

In \autoref{fig:result_combo}, setting a large number of trials for each task did not infer the best accuracy as a result. For the NC-MCI pair, using 12 trials for Fixation, Mental Imagery, Symbol Recognition tasks, or a combination of four tasks yielded acceptable results. However, using as few as 7 trials was observed to generate the best classification result, bearing up to an accuracy of 0.72 in the VERP task for the NC-MCI pair.

Acceptable results were achieved with either 12 or 17 trials during the VERP and Fixation task in the MCI-DEM pair. Using only 7 trials for the Symbol Recognition task and the 4-task combination also attained good results. The most suitable number of trials was obtained from Mental Imagery task with the minimum trials up to an accuracy of 0.75.

For NC-DEM pair, performing 7 trials in Fixation and VERP tasks or 17 trials in Symbol Recognition tasks provided admissible test outcomes. The accuracy results of the 12-trial Mental Imagery task and 17-trial 4-task combination were observed as 0.84 and 0.86, respectively. Considering the trade-offs between the results and completing more tasks with the increased number of trials, performing only Mental Imagery task was speculated to yield the best result. 

The overall outcome, as summarized in \autoref{tab:summary data}, suggested that in order to classify the cognitive declination with the proposed tasks, the ideal number of trials was less than 20 trials (17 trials + 3 testing trials) to produce the optimal performance.
The external and internal factors with the possibility to affect the performance could be related to tiredness, boredom, and drowsiness after a certain period while completing the tasks.

\begin{table}[]
\centering

\caption{\textcolor{black}{Summary of optimal number of trials, task, and Accuracy $\pm$ SE for each classification pair.}}
\renewcommand{\arraystretch}{1.2}
\begin{tabular}{l c c c c} 
\toprule[0.2em]
\textbf{Pair} & \textbf{Trials} & \textbf{Task} & \textbf{Accuracy} \\
\midrule[0.1em]
NC-MCI & 7 & VERP & 0.72 $\pm$ 0.05 \\
NC-DEM & 12 & MI & 0.84 $\pm$ 0.03 \\
MCI-DEM & 7 & MI & 0.75 $\pm$ 0.05 \\
\bottomrule[0.2em]
\end{tabular}
\label{tab:summary data}
\end{table}

\section{Discussion and Conclusion}
\label{sec_discussion}
Here, we discuss and conclude the findings from this pilot study following two proposed techniques: Relative Power-FBCSP-SVM and ERPs-PCA-SVM.

\subsection{ERPs-PCA-SVM}
The results reveal a slight difference in the event-related potentials (ERPs) \cite{wkglrm2017} from the Fixation and Mental Imagery tasks in statistical analysis. On the other hand, the Symbol Recognition and VERP tasks demonstrate the significant difference of ERPs as labeled in gray bar intervals (Figure\autoref{subfig:task11_Pz},\autoref{subfig:task12_Pz}). ERPs are visually distinguishable between the groups of participants. We can observe the most prominent amplitudes of ERPs from the NC group, followed by MCI and DEM groups. 
\textcolor{black}{
ERPs come with giant positive peaks, recognized as the P300 component. We found the declined P300 amplitude on MCI and DEM patients, corresponding to the review studies \cite{dkcokck2021,hscssk2018,morrison2019visual}. They also concluded that P300 is a sensitive biomarker for MCI and DEM diagnosis. 
}
Having ERPs as the inputs and principal component analysis (PCA) as the feature extraction, the Symbol Recognition task illustrates high classification performance as shown in \autoref{tab:erp_classification_result}. The VERP task is subordinate, while the rest do not show acceptable classification results. However, the 4-task combination provides the best classification performance, especially on NC-DEM classification.

\subsection{Relative Power-FBCSP-SVM}
In statistical analysis, all tasks provide significant differences of relative powers in EEG frequency bands as shown in \autoref{fig:tasks_stat_frequency_domain}.
In accordance with the report of previous works \cite{dkcokck2021, ltmsblyobtt2020, azjkv2013}, relative powers of dementia in the slow moving frequency bands (delta and theta) are higher than the other two groups, and are opposite in the fast moving bands (alpha and beta).
These findings motivate us to thoroughly examine an algorithm capability of extracting valuable features from each frequency band. In other words, we need a filter bank in the algorithm to divide the original signal into multiscale signals--which contain the significant frequencies of interest.
We select FBCSP as it demonstrated the impressive results in motor imagery classification for Brain-Computer Interface application. Considering ERPs, Mental Imagery seems to be the worst task in which all signals of the three groups are rarely significantly different.
On the contrary, Mental Imagery achieves the best scores for both NC-DEM and MCI-DEM classifications with FBCSP. This case emphasizes the advantage of FBCSP as another crucial feature extraction for EEG analysis in DEM recognition. The classification result of proper feature extraction and classifier is required to assess the EEG responses from the cognitive tasks. Since using SVM with features from FBCSP achieves the acceptable classification scores of all pairs, it can be one of the propitious methods to determine task efficiency.

To further validate the robustness of our method and explore relationships between Thai Mini-Mental Status Examination scores (TMSE) and the classification performance, we also tested those subjects that have equal TMSE scores.
For each classification pair, two subjects with equal TMSE scores from each class were excluded for testing, leaving the rest to be trained.
The sensitivity obtained is up to 0.82 and specificity 0.80 in NC-MCI, sensitivity 0.94 and specificity 0.91 in NC-DEM, sensitivity 0.80, and specificity 0.79 in MCI-DEM pair.
Thus, the proposed method with the practical tasks appeared to recognize the cognitively impaired subjects with similar scores.

\subsection{Contributions, Limitations and Future Directions}

To summarize, the key contributions of this work are:
(1) We investigated four visual-based cognitive tasks involving various cognitive functions: attention, working memory, and executive function. We found that different tasks are suitable for the classification of different disorders.
(2) FBCSP is one of the most effective feature extraction methods used in Brain-Computer Interfaces \cite{aczg2008, camlktmmw2020}. However, it has not been investigated on the classification of neurocognitive disorders. This is the first study that evaluates the FBCSP algorithm's feasibility to extract EEG features for dementia classification.
(3) We found that using late trials of cognitive tasks can lead to poor classification results. Since patients have mental fatigue, they might have impractical neural signals. Moreover, we determined the suitable number of trials that yield the best classification performance for each pair.
(4) To validate the effectiveness of the proposed approach, we trained and tested our model by treating subjects independently.
Classification results were reported in such a way that training, validating, and testing subjects were exclusively separated.

On the other hand, there are some limitations of this work which have to be further improved.
(1) The low number of subjects might lead to unreliable statistical analysis and classification results. Moreover, it restricts our classifier choices. We cannot use more complex classifiers such as deep neural networks because the amount of training data is insufficient. Hence, we cannot deal with the features that are exceptionally nonlinearly separated.
(2) Dementia syndrome represents a group of symptoms. Therefore, different dementia patients probably have different symptoms and different neural activity hallmarks. We do not divide the dementia class into its subclass by a symptom, resulting in degrading classifier generalization. For example, the test set contains symptoms that which training set does not have. Our classifiers might not capture neural features of these symptoms, resulting in misclassifying the patients who have these symptoms.

In future works, we need to investigate the neural activities of dementia subcategories further to eliminate the limitation. We also plan to investigate deep learning models and appropriate inductive biases to classify neurocognitive disorders to improve classification performance.

\bibliographystyle{IEEEtran}
\bibliography{Reference}

\end{document}